\documentclass[a4paper,twoside]{article}

\usepackage{epsfig}
\usepackage{subcaption}
\usepackage{calc}
\usepackage{amssymb}
\usepackage{amstext}
\usepackage{amsmath}
\usepackage{amsthm}
\usepackage{multicol}
\usepackage{pslatex}
\usepackage{apalike}
\usepackage{algorithm2e}
\usepackage[bottom]{footmisc}
\usepackage[T1]{fontenc}
\usepackage{SCITEPRESS}     

\begin{document}

\title{GAIus: Combining Genai with Legal Clauses Retrieval for Knowledge-based Assistant}

\author{\authorname{Michał Matak\sup{1}\orcidAuthor{0009-0001-3755-7158}, Jarosław A. Chudziak\sup{1}\orcidAuthor{0000-0003-4534-8652}}
\affiliation{\sup{1}Faculty of Electronics
and Information Technology, Warsaw University of Technology}
\email{\{michal.matak.stud, jaroslaw.chudziak\}@pw.edu.pl}
}


\abstract{
In this paper we discuss the capability of large language models to base their answer and provide proper references when dealing with legal matters of non-english and non-chinese speaking country.
We discuss the history of legal information retrieval, the difference between case law and statute law, its impact on the legal tasks and analyze the latest research in this field.
Basing on that background we introduce gAIus, the architecture of the cognitive LLM-based agent, whose responses are based on the knowledge retrieved from certain legal act, which is Polish Civil Code. We propose a retrieval mechanism which is more explainable, human-friendly and achieves better results than embedding-based approaches.
To evaluate our method we create special dataset based on single-choice questions from entrance exams
for law apprenticeships conducted in Poland.
The proposed architecture critically leveraged the abilities of used large language models, improving the gpt-3.5-turbo-0125 by 419\%, allowing it to beat gpt-4o and lifting gpt-4o-mini score from 31\% to 86\%.   
At the end of our paper we show the possible future path of research and potential applications of our findings.
}

\onecolumn \maketitle \normalsize \setcounter{footnote}{0} \vfill

\section{\uppercase{Introduction}}
\label{sec:introduction}

The appearance of large language models in recent years has unlocked new possibilities in natural language processing, transforming the field. These advances have significantly improved many NLP tasks and have allowed one to perform them with one single model \cite{raffel2023exploringlimitstransferlearning}. With the appearance of more advanced models they came into the mainstream and become a subject of public discussion due to their near human-like capabilities.

One of the most promising fields for recent advancements is the legal industry, where tasks often involve reading large volumes of documents, extracting key information from them, and writing substantial, repetitive amounts of text in specific jargon.

This fact, along with the aforementioned progress, raises hopes for the automation of certain tasks. The demand for such a solution remains high, as governments constantly struggle with lengthy court proceedings \cite{RPO2022LongWait,InterimResolution2000}.

A common process in the legal industry is legal research, which involves identifying and obtaining the information necessary to resolve a legal issue.
Depending on the country, this process will differ. Two main legal systems in the world may be distinguished. The first, the common law system, is characterized by the stability of the law. Court rulings in previous similar cases play a significant role. It is mainly present in Anglo-Saxon countries.
The second is the statutory law system, where current legal provisions, which may frequently change, are of greater importance. This system dominates in countries of continental Europe and countries whose legal customs originated from them, such as Mexico or Japan.

In countries with a statutory law system, it is important to address the search for relevant regulations in documents that are key sources of law (as opposed to case law), such as statutes, regulations, constitutions, and similar legal texts, due to their significant role. 

Automation in this area may not seem beneficial to experienced lawyers, but it will serve two important purposes.

The first, significant from a societal perspective, is the democratization of law, ensuring better understanding and access to legal information for all interested individuals who are not experts in the field and do not have time to read hundreds of pages that are difficult to understand.

The second, more technical purpose, is the potential to integrate such a system into more advanced architectures (e.g., multi-agent systems)  capable of providing basic legal advice or even issuing rulings in cases of low complexity. Such systems were successfully implemented for agile software engineering \cite{Chudziak2024LLM,qian2024chatdevcommunicativeagentssoftware} and software product management \cite{Cinkusz2024Agents}.
    
Our goal in this work is to develop a system that effectively enriches an AI agent with knowledge of the Polish Civil Code and to establish an evaluation process in this field. The resulting assistant will be capable of answering legal questions by referencing the appropriate provisions of the Code.

The key contributions of this paper are the architecture of the system that significantly boosts the legal knowledge of a large language model, legal act segmentation and representation method, along with a retrieval approach dispensing embeddings. Additionally, we created the benchmark based on state exams conducted in Poland, which enables a direct comparison between the performance of large language models and human capabilities, and we test currently available LLMs accordingly. 

In the second part of the article, we review the current state of research, especially concentrating on different methods of augmenting large language model with domain-specific knowledge, and the idea of AI agent. The third section details our proposed solution. In the fourth section, we outline the experiments datasets and methodology along with the results. Last but not least, the fifth section offers our conclusions, summarizes the paper, and explores potential directions for future work.

\section{\uppercase{Background}}

To outline the background for our discussion, we focus on three key areas: methods of augmenting large language models, challenges and improvements in legal information retrieval, and legal assistants evaluation methods.

\subsection{Augmenting Large Language Models}
There are several methods to augment LLMs to enhance it with domain-specific knowledge and provide better responses. 
One of the most traditional methods is fine-tuning, which involves training LLM with data specific to a given field of knowledge. This method was proved to be efficient in several cases, such as finance \cite{Jeong_2024}.

The use of fine-tuning to teach a model legal regulations can be quite problematic. Legal regulations are often subject to frequent changes, including the addition of new rules, modifications, and the removal of old ones. Due to the large number of legislative changes, if a large number of legal acts were incorporated, fine-tuning would need to occur frequently — potentially every few days — to ensure up-to-date performance. This requirement makes fine-tuning both resource-intensive and difficult to maintain over time.

An alternative method to fine-tuning is Retrieval-Augmented Generation (RAG) proposed by \cite{lewis2021retrievalaugmentedgenerationknowledgeintensivenlp}. The authors introduced a model with types of memory: a parametric memory, implemented as a pre-trained seq2seq transformer, and a non-parametric memory, which encodes documents and queries using pre-trained encoder models. The retrieval process selects the top $k$ documents by maximizing the inner product between their encoded forms and the encoded query.
The method achieved state-of-the-art results on open Natural Questions, WebQuestions, and CuratedTrec and strongly outperform recent approaches that use specialized pre-training objectives on TriviaQA.

Research shows that RAG leads to improvements that are comparable to those achieved by fine-tuning \cite{balaguer2024ragvsfinetuningpipelines}. Additionally, RAG requires lower costs at the initial stage; however, it might incur higher costs in the long run due to increased token usage \cite{balaguer2024ragvsfinetuningpipelines}.

To solve more complex tasks, LLMs were integrated with other tools such as internet search, memory, planning, and external tool usage. In these architectures, the role of LLM is limited not only to return the proper answer, but also to break the task into subtasks and orchestrate the use of different tools to achieve the desired result. 

A straightforward example of this concept is ReAct introduced in \cite{yao2023reactsynergizingreasoningacting}. It is a reasoning framework where LLM can either perform actions on external environment or perform actions in language space that lead to the generation of thoughts that are aimed to compose useful information to improve performing reasoning in current context. On HotpotQA and Fever it achieved better results that standard LLM, Act framework and chain-of-thoughts-prompting. 

The landscape of agents and large language models was structured in \cite{sumers2024cognitivearchitectureslanguageagents}, where three uses of large language models from agent system perspective were distinguished: producing output from input (raw LLM), a language agent in a direct feedback loop with the environment, and a cognitive language agent that manages its internal state through learning and reasoning. 
The paper further explores different architectures of cognitive agents and organizes them in three dimensions: information storage, action space, and decision-making procedure.

\subsection{Legal Information Retrieval}
When retrieving information for legal issues, there are several challenges worth considering to understand the specific requirements of this field. Some of them are exceptionally relevant in the context of using LLMs for this task.

First, when obtaining provisions from legal acts, it is crucial to accurately convey their content. In many cases, subtle details - such as punctuation marks, as well as the difference between "among others" and "in particular" - can be extremely important. In certain cases, even the placement of a provision within a statute can be significant. When using fine-tuned LLMs, there is a substantial risk of returning imperfect answers. This risk arises from the potential for hallucinations (i.e., generating plausible but incorrect content) and the difficulty of verifying responses without specialized legal knowledge.

Second, the goal in legal research is to find all relevant information on a given topic, not just a portion of it. Considering that in most legal systems, a specific provision overrides a general one, omitting it may result in a completely erroneous legal interpretation.

Finally, it is also important to consider the customs adopted in different countries when using base models. Unlike in fields such as medicine or finance, there is no universal "objective truth" in law. Legal regulations can vary widely across jurisdictions, and in some cases, regulatory approaches may even be completely opposite.

One of the first approaches to align LLMs with the legal domain was LegalBert introduced in \cite{chalkidis2020legalbertmuppetsstraightlaw}. In this study, two variants based on BERT architecture were proposed: one was BERT model further pre-trained on legal corpora and the second one was BERT model pre-trained from the scratch on this corpus. LegalBert achieved state-of-the-art results on three tasks (multi-label classification in ECHR-CASES and
contract header, lease details in CONTRACTS-NER).

A similar approach was used in LawGPT \cite{zhou2024lawgptchineselegalknowledgeenhanced} that was designed for Chinese legal applications. In addition to legal domain pre-training on a corpus consisting of 500K legal documents,the  authors proposed the step of Legal-supervised fine-tuning (LFT) to enhance the results. The model achieved better results than open source models in 5 out of 8 cases; however, it still largely falls behind the GPT-3.5 Turbo and GPT-4 models.

The platform that fully integrated large language models and retrieval-augmented generation was CaseGPT \cite{yang2024casegptcasereasoningframework}, a system designed for case-based reasoning in the legal and medical domains. It consisted of three modules: a query processing module that transformed user queries for retrieval, a case retrieval engine based on a dense vector index with a semantic search algorithm, and an insight generation module that analyzes retrieved cases within the user's query context.

A different approach to case retrieval was presented in \cite{wiratunga2024cbrragcasebasedreasoningretrieval}. In this approach, each case consists of a question, an answer, supporting evidence for the answer, and the named entities extracted from this support. For retrieval, the authors proposed dual embeddings: intra-embeddings for case query encoding and inter-embeddings for encoding support and entities. The evaluation demonstrated that this architecture improves performance by 1.94\% compared to the baseline without RAG.

A notable solution in legal information retrieval for statute law was the winning solution for Task 3 in the Competition on Legal Information Extraction/Entailment 2023 (COLIEE 2023), developed by the CAPTAIN team \cite{nguyen2024captaincoliee2023efficient}. Its purpose was to extract a subset of articles of the Japanese Civil Code from the entire Civil Code to answer questions from Japanese legal bar exams. The solution was based on negative sampling, multiple model checkpoints (with the assumption that each checkpoint is biased toward specific article categories), and a specific training process that included a data-filtering approach.

Another study conducted on the COLIEE 2023 dataset proposes the use of prompting techniques in the final stage of the retrieval system, preceded by BM25 pre-ranking and BERT-based re-ranking \cite{nguyen2024enhancinglegaldocumentretrieval}. The solution in this study achieved a score that outperformed the aforementioned approach (the solution of the winning team) by 4\%.

\subsection{Legal Assistants Evaluation Methods}
There are several datasets for evaluating legal agents. One of the most notable is LegalBench \cite{guha2023legalbenchcollaborativelybuiltbenchmark}, which consists of 162 tasks designed and handcrafted by legal professionals. The tasks cover six distinct types of legal reasoning. This benchmark inspired LegalBench-RAG \cite{pipitone2024legalbenchragbenchmarkretrievalaugmentedgeneration}, a dataset tailored specifically for retrieval-augmented generation in the legal domain. LegalBench-RAG includes 6,858 query-answer pairs over a corpus containing more than 79 million characters, created from documents sourced from four distinct datasets.

For the task of Extreme Multi-Label Legal Text Classification \cite{chalkidis-etal-2019-extreme}, a dataset was created from 57,000 legislative documents in Eurlex—the public document database of the European Union. These documents were annotated with EUROVOC, a multidisciplinary thesaurus used to standardize legal terminology across various domains.

Finally, an interesting evaluation benchmark was used for Task 3 of the 2023 Competition on Legal Information Extraction/Entailment (COLIEE 2023). The dataset consists of questions that require the evaluated agent to answer yes/no and to provide the relevant articles of the Japanese Civil Code. Notably, this dataset is one of the very few examples that focuses specifically on the Statute Law Retrieval Task.

\section{\uppercase{Assistant Architecture}}  
The architecture of gAIus consists of two key components. The first is the query flow, which defines how the agent interacts with human input and the available tools. The second is the retrieval mechanism, responsible for selecting documents relevant to the given query.  

\subsection{Query Flow}
\begin{figure}[h]
  \centering
   {\epsfig{file = 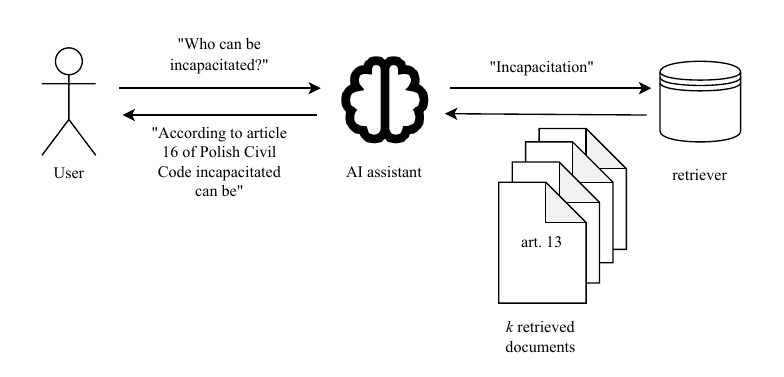, width = \linewidth}}
  \caption{Visualisation of query flow.}
  \label{fig:flow}
 \end{figure}

The flow in our architecture illustrated in Figure \ref{fig:flow}. 

The architecture consists of a single agent that receives a query from the user and interacts with a single tool: the retriever. Upon receiving the request, the agent reformulates the query to retrieve relevant documents. The purpose of this architecture is to enable the LLM to extract the semantically important part of the question and generate a more general retrieval query.

For example, a specific question such as "Who can be incapacitated?" would be transformed into the more general query "Incapacitation." By generalizing the query, the system can increase the likelihood of retrieving the correct documents.

The agent can use the retriever multiple times, allowing it to attempt different queries if the relevant documents are not found on the first attempt. This iterative approach increases the chance of successfully retrieving the correct information.

\subsection{Retrieval}
To provide context for a large language model (LLM), one approach is to include the entire document (e.g., a legal act) in the model's context. However, there are several challenges associated with this approach. First, the document length may be too long and simply does not fit within the LLM's context window. Second, LLMs tend to perform worse when processing long contexts, often failing to extract the necessary information. Third, most commercial LLMs are billed based on token usage, so including the entire document in the context would result in significantly increased costs.

For this reason, we decided to chunk the Polish Civil Code before storing it in the document database. The chunking method can vary and significantly influence the quality of the retrieval process.

\begin{figure}[h]
  \centering
   {\epsfig{file = 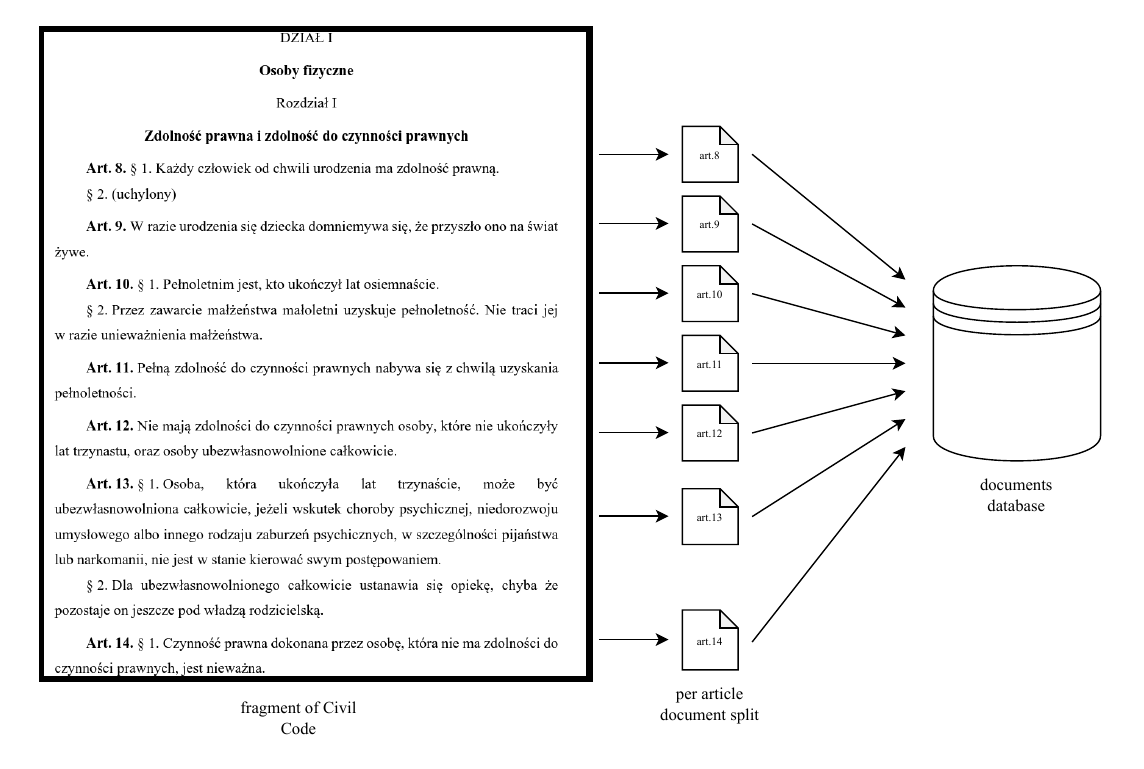, width = \linewidth}}
  \caption{Visualisation of Civil Code chunking for retrieval.}
  \label{fig:chunking}
 \end{figure}

The Polish Civil Code is already divided by the lawmaker into various entities, such as books, titles, chapters, and sections. However, the main editorial unit, which consists of one or more sentences, is the article. We chose to split the entire Code into individual articles, which we will further refer to as documents. The chunking method is illustrated in Figure \ref{fig:chunking}.

To assess the relevance of the document to the query, we score the documents using a custom function and retrieve the top-$k$ documents with the highest scores (Algorithm \ref{alg:document} and \ref{alg:part}).

\begin{algorithm}[h]
 \caption{Document scoring function.}
  \KwData{document, query}
 \KwResult{score of the document}
 $max\_score \gets 0$ \;
 \For{$i \gets 0$ to $|document| - |query|$}{
  $part \gets document[i : i + |query|]$\;
  $part\_score \gets \textsc{ScorePart}(part, query)$\;
  \If{part\_score > max\_score}{
   $max\_score \gets part\_score$\;
   }{
  }
 }
   \label{alg:document}
\end{algorithm}
\begin{algorithm}[h]
 \caption{Part scoring function.}
  \KwData{part, query}
 \KwResult{score of the part }
    $score \gets 0$ \;
    \For{$i \gets 0$ to $|part| - 1$}{
        \If{$part[i] = query[i]$}{
            $score \gets score + 1$
        }
    }
      \label{alg:part}
\end{algorithm}

The algorithm's behavior can be described as calculating the number of matching letters between the query and each part of the document, constrained by the query length. This fuzzy text search approach is inspired by the behavior of law students when searching for regulations. Typically, they visually scan the text, focusing on target words to identify the required regulation. Additionally, this method leverages existing LLM knowledge, as models often have some understanding of legal texts and are able to cite relevant portions of regulations.

To fully realize the potential of the retrieval process, we utilize the earlier decision to split the legal code into articles. All chunks are relatively small, with an average length of approximately 140 tokens.

For this reason, we chose to retrieve a relatively large number of chunks ($k=50$). The cognitive capabilities of the large language model allow it to filter out irrelevant articles, ensuring that the pipeline functions correctly as long as the retrieval achieves satisfactory recall. In the worst-case scenario, where the longest chunks are retrieved, no more than 30 chunks fit within the context window of smaller models (e.g., GPT-3.5-turbo). Although this could theoretically cause the pipeline to fail, our experiments indicate that this risk is relatively low.

To differentiate the impact of the document scoring method and retrieval parameters (including code segmentation), we also implemented the architecture using embeddings. We refer to this variant as gAIusRAG.

\subsection{Used Technologies}
The entire project is written in Python. To create the assistant, we used the LangChain framework with its agents API \cite{LangChainAgents2024}. We implemented a custom retriever based on the LangChain retriever interface. For the vectorstore, we used ChromaDB, and for Polish language embeddings, we employed mmlw-roberta-large encodings from \cite{polish-nlp-resources}.

The Polish Civil Code was downloaded from the official government website in PDF format \cite{isap1964kodekscywilny}. The document was subsequently converted into a TXT file and preprocessed by removing footnotes, publication metadata, and annotations.

\section{\uppercase{Experiments}}
In the experiments part of the article we would like to discuss the dataset we created for evaluation purposes, the methodology and the results obtained.

\subsection{Dataset}
We created the evaluation dataset similar to the one used on COLIEE 2023 competition. It consists from questions from entrance exams for law apprenticeships from the years 2021-2023. 
These exams are state exams conducted every year in Poland. Law graduates have to pass them to enroll for a 3 years apprenticeship program. After its completion and another exam (counterpart of bar exam) they become law professionals. 
Every year three such exams are conducted: for notarial apprenticeship, for bailiff apprenticeship, and  for attorney and legal advisers' apprenticeship.

Each of the exams consists of 150 closed questions with three possible answers: a, b, or c. To pass the test, one has to answer at least 100 correctly. The questions are usually quite simple if one knows the proper regulation and they do not need extensive reasoning. Hence comes our decision to use them as an evaluation framework for our assistant.
The questions were downloaded from the government website \cite{gov2023examquestions} along with the answers. The answers file also contained the indices of law regulations needed to answer each question.
After filtering out all questions that did not referred to the Civil Code the dataset consisted of 146 different questions.

\subsection{Methodology}
We compare our architecture, powered by GPT-3.5-turbo-0125 and GPT-4o-mini models, with other OpenAI LLMs without any additional enhancements. All models were configured with a temperature of 00 to ensure reproducibility. The evaluation was performed on RAG variants to compare them with our retrieval method.

The primary goal of the proposed law assistant is to provide accurate and relevant information about various legal issues. For this reason, during the evaluation process, we do not only expect the assistant to return the correct answer but also to justify its response and cite the relevant legal article.

To achieve this, each evaluated assistant was set up with the following system prompt:
\begin{small}
\begin{verbatim}
You are a helpful assistant specializing 
in Polish law. You will receive questions from 
an exam, each consisting of a question or an 
incomplete sentence followed by three possible
answers labeled a, b, and c. 

Your task is to:
1. Choose the correct answer.
2. Provide a detailed explanation for 
your choice.
3. Refer to the relevant article(s) 
in the one of polish regulations.

Please ensure your responses are precise and 
informative. Respond in polish.
\end{verbatim}
\end{small}

When assessing the performance of the assistant, we check if the correct answer was chosen and if appropriate regulation was referred to. 
We measure the number of successfully answered questions (answer score), the number of correctly referred contexts (context score), and the number of situations in which both answer and context are correct (joint score). If the assistant refers to more regulations than necessary, we still consider it a good answer, as long as the extra references are minimal (typically no more than two articles).

To automate the evaluation, we developed a simple evaluation agent that extracts the chosen answer and the cited articles into a structured JSON format. Based on these extracted values, the assistant response is evaluated.

\begin{table*}[ht]
\caption{Example question from evaluation dataset and assistants responses (originally in Polish, translated to English).}\label{tab:sample_question} \centering
\begin{tabular}{|p{22mm}|p{128mm}|}
\hline
Question 
 & According to the Civil Code, the commercial power of attorney (prokura):

a) can always be limited with effect against third parties;

b) can never be limited with effect against third parties;

c) cannot be limited with effect against third parties, unless a specific provision states otherwise. \\ \hline

Correct answer & C \\ \hline

Correct context &  Art. $109^1.$ § 2. The commercial power of attorney (prokura) cannot be limited with effect against third parties, unless a specific provision states otherwise. \\ 
\hline

gAIus answer & The correct answer is:

c) cannot be limited with effect against third parties, unless a specific provision states otherwise.

According to Article 1091 § 2 of the Civil Code, a commercial power of attorney (prokura) cannot be limited with effect against third parties unless a specific legal provision allows such a limitation. This means that a power of attorney granted by an entrepreneur cannot be restricted with respect to third parties, unless there is a specific legal regulation permitting such a restriction. This is crucial for the protection of third parties who may rely on the agent's authority in dealings with the entrepreneur.

Legal provision: Article 1091 § 2 of the Civil Code. \\ \hline

gpt-4o answer &
Answer: b) can never be limited with effect against third parties.

Explanation: According to the Civil Code, the commercial power of attorney (prokura) is a special type of power of attorney granted by an entrepreneur who is subject to registration in the business register. Prokura includes authorization for judicial and extrajudicial actions related to running the business. According to Article $109^3$ § 2 of the Civil Code, the limitation of prokura is not effective against third parties. This means that any limitations on the scope of prokura are effective only internally, between the entrepreneur and the agent, but not against third parties.

Legal basis: Article $109^3$ § 2 of the Civil Code: "The commercial power of attorney cannot be limited with effect against third parties." \\ \hline
\end{tabular}
\end{table*}

\subsection{Results}
To present the detailed behavior of the assistants we show and discuss sample question from dataset with responses from gpt-4o and gAIus based on gpt-3.5-turbo in \ref{tab:sample_question}. 

In this table, a sample question from the evaluation dataset is shown, alongside the correct answer, the correct context, and the responses from the two evaluated assistants. The original source information was in Polish, but was translated into English for the reader’s convenience. The correct context clearly indicates the proper answer to the question.

Our assistant, gAIus, correctly identifies and references the relevant article (although the index number became part of the ID due to preprocessing). As a result, gAIus provides the correct answer. In contrast, GPT-4o chooses the wrong answer and hallucinates a context that appears similar to the desired one. In particular, in the Polish Civil Code, there is no paragraph $109^3$ §2, and the referenced text lacks the key information necessary to answer the question.

The evaluation results for all questions from the dataset are presented in Table \ref{tab:results}, with several notable outcomes:

At first, ll raw commercial large language models performed better in the answer score than in the context score. Among these models, except for GPT-4o, a significant disparity between the answer score and the context score is observed, with the answer score prevailing. In contrast, the gAIus assistants excelled in providing the correct context. GPT-4o emerged as the best-performing commercial model. However, its answer score ranked third compared to all evaluated models.

Secondly, the gAIus architecture significantly enhanced the performance of the underlying LLMs. In the case of
gpt-3.5-turbo-0125, the joint score improved by 419\% (from 21 to 109) achieving better results, outperforming the best commercial model, GPT-4o. The results of gpt-4o-mini improved by 173\% and gAIus, with gAIus achieving the highest joint score and context score, and the second-best answer score.

Furthermore, architectures based on our scoring function outperformed those using RAG. However, the RAG variant based on GPT-4o-mini achieved the best answer score, which was slightly higher than that of the corresponding gAIus variant using search functions.

\begin{table*}[!ht]
\caption{Results of evaluation on 146 questions.}\label{tab:results} \centering
\begin{tabular}{lccc}
  \hline
  Assistant & Answer score & Context score & Joint score\\
  \hline
  gpt-3.5-turbo-0125 & 77 & 26 & 21\\
  gpt-4o-mini & 101 & 58 & 46\\
  gpt-4 & 115 & 53 & 44\\
  gpt-4o & 127 & 106 & 98\\ \hline
  gAIusRAG based on gpt-3.5-turbo-0125 & 101 & 116 & 91\\
  gAIusRAG based on gpt-4o-mini & \textbf{130} & 128 & 117\\
  gAIus based on gpt-3.5-turbo-0125& 112 & 126 & 109\\
  gAIus based on gpt-4o-mini& 128 & \textbf{139} & \textbf{126}\\
  \hline
\end{tabular}
\end{table*}

\section{\uppercase{Conclusions, Summary and Further work}}
Several possible developments can be identified for our research:

First, the legal system in Poland has sources beyond the Civil Code. A natural path for further development would be to expand the assistant to cover additional legal acts. This could be achieved by either integrating an additional retrieval tool for the single-agent system or designing a multi-agent architecture with a routing agent to manage queries across multiple sources.

Second, another significant enhancement would be the addition of case retrieval relevant to the legal issue at hand. Although case law plays a limited role in statutory law systems, it still serves as a powerful instrument for argumentation, particularly when cases originate from renowned courts.

Finally, our solution can serve as the foundation for an AI system capable of solving more complex legal cases. The questions in our current evaluation framework were relatively simple and focused on direct queries. It would be valuable to examine how large language models perform in a more realistic legal environment, where reasoning and multi-step analysis are required.

\section*{\uppercase{Acknowledgements}}
The translation (from Polish to English) of the example dataset question with the answers and other cited regulations was performed using the ChatGPT-4o model \cite{chatgpt2023}. 

\bibliographystyle{apalike}
{\small
\bibliography{gaius}}

\end{document}